\documentclass[11pt,a4paper]{article}
\usepackage[hyperref]{emnlp-ijcnlp-2019}
\usepackage{times}
\usepackage{latexsym}

\usepackage{url}

\usepackage{hyperref}
\usepackage{url}
\usepackage{subcaption} 
\usepackage{booktabs} %
\usepackage{enumitem}
\usepackage{graphicx}
\usepackage{bbm}
\usepackage{amsmath,amsfonts,bm}

\setlist[enumerate,1]{leftmargin=1.2em,labelindent=0em,itemsep=0pt,labelsep*=0.5em}
\aclfinalcopy %

\title{Weakly Supervised Attention Networks \\for Fine-Grained Opinion Mining and Public Health}

\author{Giannis Karamanolakis, Daniel Hsu, Luis Gravano\\
Columbia University, New York, NY 10027, USA \\
\texttt{\{gkaraman, djhsu, gravano\}@cs.columbia.edu}
}

\date{}

\begin{document}
\maketitle
\begin{abstract}
In many review classification applications, a fine-grained analysis of the reviews is desirable, because different segments  (e.g., sentences) of a review may focus on different aspects of the entity in question. 
However, training supervised models for segment-level classification requires segment labels, which may be more difficult or expensive to obtain than review labels. 
In this paper, we employ Multiple Instance Learning (MIL) and use only weak supervision in the form of a single label per review.
First, we show that when inappropriate MIL aggregation functions are used, then MIL-based networks are outperformed by simpler baselines.
Second, we propose a new aggregation function based on the sigmoid attention mechanism and show that our proposed model outperforms the state-of-the-art models for segment-level sentiment classification (by up to 9.8\% in F1). 
Finally, we highlight the importance of fine-grained predictions in an important public-health application: finding actionable reports of foodborne illness.
We show that our model achieves 48.6\% higher recall compared to previous models, thus increasing the chance of identifying previously unknown foodborne outbreaks.
\end{abstract}

\section{Introduction}

\begin{figure}[t]
\includegraphics[height = 4.8cm, width = 7.5cm]{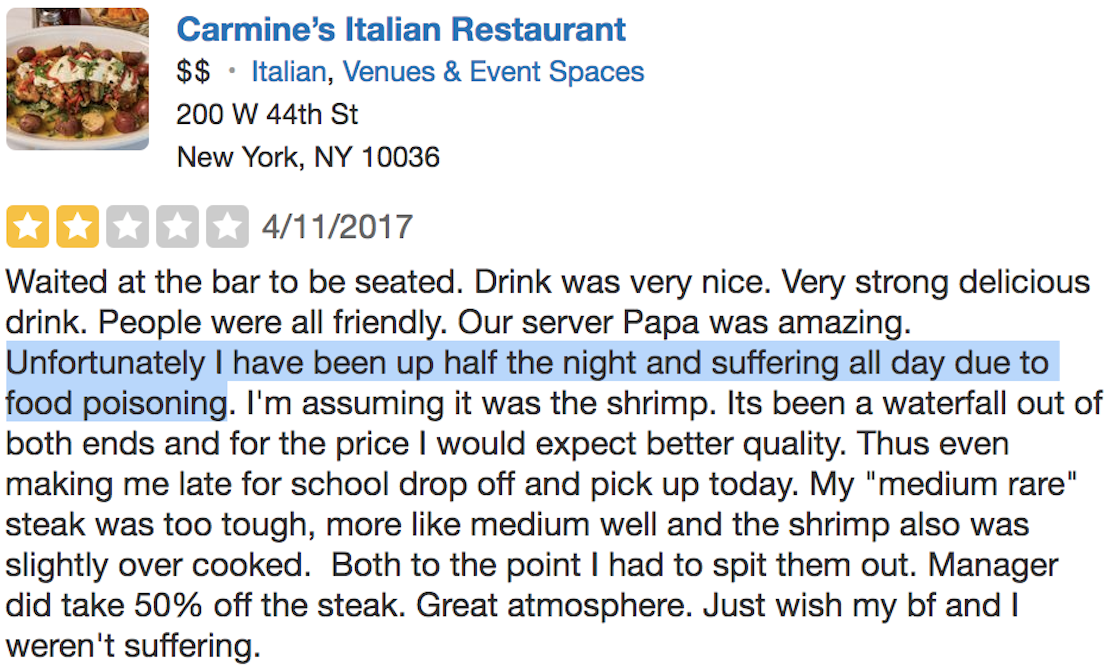}
	\caption{\small{A Yelp review discussing both positive and negative aspects of a restaurant, as well as food poisoning.}}
\label{fig:food_poisoning_example}
\end{figure}

Many applications of text review classification, such as sentiment analysis, can benefit from a fine-grained understanding of the reviews.
Consider the Yelp restaurant review in Figure~\ref{fig:food_poisoning_example}.
Some segments (e.g., sentences or clauses) of the review express positive sentiment towards some of the items consumed, service, and ambience, but other segments express a negative sentiment towards the price and food.
To capture the nuances expressed in such reviews, analyzing the reviews at the segment level is desirable.

In this paper, we focus on segment classification when only review labels---but not segment labels---are available. 
The lack of segment labels prevents the use of standard supervised learning approaches.
While review labels, such as user-provided %
ratings, are often available, they are not directly relevant for segment classification, thus presenting a challenge for supervised learning.
Existing weakly supervised learning frameworks have been proposed for training models such as support vector machines~\cite{andrews2003support,yessenalina2010multi,gartner2002multi}, logistic regression~\cite{kotzias2015group}, and hidden conditional random fields~\cite{tackstrom2011discovering}.
 The most recent state-of-the-art approaches employ the Multiple Instance Learning (MIL) framework (Section~\ref{s:multiple-instance-learning}) in hierarchical neural networks~\cite{pappas2014explaining,kotzias2015group,angelidis2018multiple,pappas2017explicit,ilse2018attention}. 
  MIL-based hierarchical networks combine the (unknown) segment labels through an aggregation function to form a single review label. 
  This enables the use of ground-truth review labels as a weak form of supervision for training segment-level classifiers.
However, it remains unanswered whether performance gains in current models stem from the hierarchical structure of the models or from the representational power of their deep learning components. 
Also, as we will see, the current modeling choices for the MIL aggregation function might be problematic for some applications and, in turn, might hurt the performance of the resulting classifiers. 
As a first contribution of this paper, we show that non-hierarchical, deep learning approaches for segment-level sentiment classification ---with only review-level labels--- are strong, and they equal or exceed in performance hierarchical networks with various MIL aggregation functions.

As a second contribution of this paper, we substantially improve previous hierarchical approaches for segment-level sentiment classification and propose the use of a new MIL aggregation function based on the sigmoid attention mechanism to jointly model the relative importance of each segment as a product of Bernoulli distributions.
This modeling choice allows multiple segments to contribute with different weights to the review label, which is desirable in many applications, including segment-level sentiment classification. %
We demonstrate that our MIL approach outperforms all of the alternative techniques.

As a third contribution, we experiment beyond sentiment classification and apply our approach to a critical public health application: the discovery of foodborne illness incidents in online restaurant reviews.
Restaurant patrons increasingly turn to social media---rather than official public health channels---to discuss food poisoning incidents (see Figure~\ref{fig:food_poisoning_example}).
As a result, public health authorities need to identify such rare incidents among the vast volumes of content on social media platforms.
We experimentally show that our MIL-based network effectively detects segments discussing food poisoning and has a higher chance than all previous models to identify unknown foodborne outbreaks. 
\section{Background and Problem Definition}
\label{s:background-section}
We now summarize relevant work
on fully supervised (Section~\ref{s:supervised-sentence-level-classification}) and weakly supervised models (Section~\ref{s:multiple-instance-learning}) for segment classification.
We also describe a public health application for our model evaluation (Section~\ref{s:background-foodborne-illness-discovery}). Finally, we define our problem of focus (Section~\ref{s:problem-definition}).

\subsection{Fully Supervised Models}
\label{s:supervised-sentence-level-classification}
State-of-the-art supervised learning methods for segment classification use segment embedding techniques followed by a classification model. 
During segment encoding, a segment $s_i = (x_{i1}, x_{i2}, \dotsc, x_{iN_i})$ composed of $N_i$ words is encoded as a fixed-size real vector $h_i \in \mathbb{R}^\ell$ using transformations such as the average of word embeddings~\cite{wieting2015towards,arora2016simple}, Recurrent Neural Networks (RNNs)~\cite{wieting2017revisiting,yang2016hierarchical}, Convolutional Neural Networks (CNNs)~\cite{kim2014convolutional}, or 
self-attention blocks~\cite{devlin2019bert,radford2018improving}. 
We refer to the whole segment encoding procedure as 
 $h_i = \operatorname{ENC}(s_i)$.
During segment classification, the segment $s_i$ is assigned to one of $C$ predefined classes $[C]:=\{1,2,\dotsc,C\}$.
To provide a probability distribution $p_i= \langle p_i^1, \dotsc, p_i^C \rangle$ over the $C$ classes, the segment encoding $h_i$ is 
fed to a classification model:
 $   p_i = \operatorname{CLF}(h_i)$.
Supervised approaches require ground-truth {\em segment labels} for training.

\subsection{Weakly Supervised Models}
\label{s:multiple-instance-learning}
State-of-the-art weakly supervised approaches for segment and review classification employ the Multiple Instance Learning (MIL) framework~\cite{zhou2009multi, pappas2014explaining, kotzias2015group, pappas2017explicit, angelidis2018multiple}.
In contrast to traditional supervised learning, where {\em segment labels} are required to train segment classifiers, MIL-based models can be trained using {\em review labels} as a weak source of supervision, as we describe next. 

\begin{figure}[t]
    \includegraphics[height = 5.5cm, width=1\columnwidth]{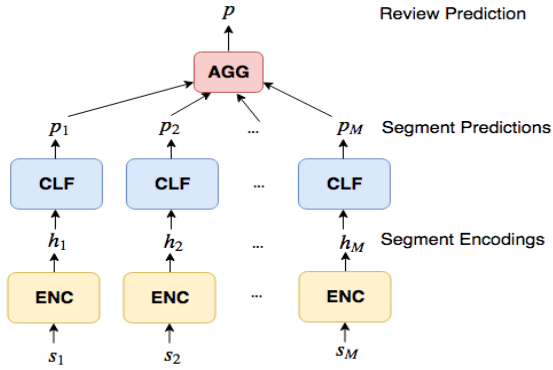}
    \caption{MIL-based hierarchical models.}
    \label{fig:weakly_supervised_segment_level_classification}
\end{figure}

MIL is employed for problems where data are arranged in groups (bags) of instances. 
In our setting, each review is a group of segments:  $r=(s_1,s_2, \dotsc,s_M)$. 
The key assumption followed by MIL is that the observed review label is an aggregation function of the unobserved segment labels: $p=\operatorname{AGG}(p_1,\dotsc,p_M)$.
Hierarchical MIL-based models (Figure~\ref{fig:weakly_supervised_segment_level_classification}) work in three main steps: (1) encode the review segments into fixed-size vectors $h_i = \operatorname{ENC}(s_i)$, 
(2) provide segment predictions $p_i = \operatorname{CLF}(h_i)$, and (3) aggregate the predictions to get a review-level probability estimate $p=\operatorname{AGG}(p_1,\dotsc,p_M)$. 
Supervision during training is provided in the form of review labels.

Different modeling choices have been taken for each part of the MIL hierarchical architecture. 
\citet{kotzias2015group} encoded sentences as the internal representations of a hierarchical CNN that was pre-trained for document-level sentiment classification~\cite{denil2014extraction}.
For sentence-level classification, they used logistic regression, while the aggregation function was the uniform average. 
\citet{pappas2014explaining,pappas2017explicit} employed Multiple Instance Regression, evaluated various models for segment encoding, including feed forward neural networks and Gated Recurrent Units (GRUs)~\cite{bahdanau2014neural}, and used the weighted average for the aggregation function, where the weights were computed by linear regression or a one-layer neural network.
\citet{angelidis2018multiple} proposed an end-to-end Multiple Instance Learning Network (MILNET), which outperformed previous models for sentiment classification using CNNs for segment encoding, a softmax layer for segment classification, and GRUs with attention~\cite{bahdanau2014neural} 
to aggregate segment predictions as a weighted average.
Our proposed model (Section~\ref{s:model-description}) also follows the MIL hierarchical structure of Figure~\ref{fig:weakly_supervised_segment_level_classification} for both sentiment classification and an important public health application, which we discuss next.

\subsection{Foodborne Illness Discovery in Online Restaurant Reviews}
\label{s:background-foodborne-illness-discovery}
Health departments nationwide have started to analyze social media content (e.g., Yelp reviews, Twitter messages) to identify foodborne illness outbreaks originating in restaurants. %
In Chicago~\cite{harris2014health}, New York City~\cite{effland2018discovering}, Nevada~\cite{sadilek2016deploying}, and St. Louis~\cite{harris2018evaluating}, text classification systems have been successfully deployed for the detection of social media documents mentioning foodborne illness. (Figure~\ref{fig:food_poisoning_example} shows a Yelp review discussing a food poisoning incident.)
After such social media documents are flagged by the classifiers, they are typically examined manually by epidemiologists, who decide if further investigation (e.g., interviewing the restaurant patrons who became ill, inspecting the restaurant) is warranted. 
This manual examination is time-consuming, and hence it is critically important to (1) produce accurate review-level classifiers, to identify foodborne illness cases while not showing epidemiologists large numbers of false-positive cases; and (2) annotate the flagged reviews to help the epidemiologists in their decision-making. 

We propose to apply our segment classification approach to this important public health application. 
By identifying which review segments discuss food poisoning, epidemiologists could focus on the relevant portions of the review and safely ignore the rest. 
As we will see, our evaluation will focus on Yelp restaurant reviews.
Discovering foodborne illness is fundamentally different from sentiment classification, because the mentions of food poisoning incidents in Yelp are rare. 
Furthermore, even reviews mentioning foodborne illness often include multiple sentences unrelated to foodborne illness (see Figure~\ref{fig:food_poisoning_example}).

\subsection{Problem Definition}
\label{s:problem-definition}
Consider a text review for an entity, with $M$ contiguous segments $r=(s_1, \ldots, s_M)$.
Segments may have a variable number of words and different reviews may have a different number of segments. 
A discrete label $y_r \in [C]$ is provided for each review but the individual segment labels are not provided.
Our goal is to train a segment-level classifier that, given an unseen test review $r^t=(s^t_1,s^t_2, \dotsc,s^t_{M_t})$, predicts a label $p_i \in [C]$ for each segment and then aggregates the segment labels to infer the review label $y_r^t \in [C]$ for $r^t$.

\section{Non-Hierarchical Baselines}
\label{s:non-hierarchical-baselines}
We can address the problem described in Section~\ref{s:problem-definition} without using hierarchical approaches such as MIL.
In fact, the hierarchical structure of Figure~\ref{fig:weakly_supervised_segment_level_classification} for the MIL-based deep networks adds a level of complexity that has not been empirically justified, giving rise to the following question: do performance gains in current MIL-based models stem from their hierarchical structure or just from the representational power of their deep learning components?

We explore this question by evaluating a class of simpler non-hierarchical baselines: deep neural networks trained at the {\em review level} (without encoding and classifying individual segments) and applied at the {\em segment level} by treating each test segment as if it were a short ``review.'' 
While the distribution of input length is different during training and testing, we will show that this class of non-hierarchical models is quite competitive and sometime outperforms MIL-based networks with inappropriate modeling choices. 

\section{Hierarchical Sigmoid Attention Networks}
\label{s:model-description}
We now describe the details of our MIL-based hierarchical approach, which we call Hierarchical Sigmoid Attention Network (HSAN). 
HSAN works in three steps to process a review, following the general architecture in Figure~\ref{fig:weakly_supervised_segment_level_classification}:
(1) each segment $s_i$ in the review is encoded as a fixed-size vector using word embeddings and CNNs~\cite{kim2014convolutional}: $h_i = \operatorname{CNN}(s_i) \in \mathbb{R}^\ell$;
(2) each segment encoding $h_i$ is classified using a softmax classifier with parameters $W\in \mathbb{R}^{\ell}$ and $b \in \mathbb{R}$: 
$p_i = \operatorname{softmax}(W h_i + b)$; and (3) a review prediction $p$ is computed as an aggregation function of the segment predictions $p_1,\dotsc,p_M$ from the previous step.  
A key contribution of our work is the motivation, definition, and evaluation of a suitable aggregation function for HSAN, a critical design issue for MIL-based models.

The choice of aggregation function has a substantial impact on the performance of MIL-based models and should depend on the specific assumptions about the relationship between bags and instances~\cite{carbonneau2018multiple}. 
Importantly, the performance of MIL algorithms depends on the witness rate (WR), which is defined as the proportion of positive instances in positive bags.
For example, when WR is very low (which is the case in our public health application), using the uniform average as an aggregation function in MIL is not an appropriate modeling choice, because the contribution of the few positive instances to the bag label is outweighed by that of the negative instances. 

The choice of the uniform average of segment predictions~\cite{kotzias2015group} is also problematic because particular segments of reviews might be more informative than other segments for the task at hand and thus should contribute with higher weights to the computation of the review label.
For this reason, we opt for the weighted average~\cite{pappas2014explaining,angelidis2018multiple}:
\begin{equation}
    p = \frac{\sum_{i=1}^M \alpha_i \cdot p_i}{\sum_{i=1}^M \alpha_i}.
\label{eq:weighted-avg-aggregation}
\end{equation}
The weights $\alpha_1,\dotsc,\alpha_M \in [0,1]$ define the relative contribution of the corresponding segments $s_1,\dotsc,s_M$ to the review label. 
To estimate the segment weights, we adopt the attention mechanism~\cite{bahdanau2014neural}.
In contrast to MILNET~\cite{angelidis2018multiple}, which uses the  traditional softmax attention, we propose to use the sigmoid attention. 
Sigmoid attention is both functionally and semantically different from softmax attention and is more suitable for our problem, as we show next.

The probabilistic interpretation of softmax attention %
is that of a categorical latent variable $z \in \{1, \dotsc , M\}$ that represents the index of the segment to be selected from the $M$ segments~\cite{kim2017structured}.  
The attention probability distribution is:
\begin{equation}
    p(z = i \mid e_1, \dotsc, e_M) =  \frac{\exp(e_i)}{\sum_{i=1}^M \exp(e_i)}, 
  \label{eq:softmax-attention-weights}
\end{equation}
 where:
 \begin{equation}
    e_i = u_a^T \tanh(W_a h_i' + b_a),
\label{eq:attention-similarity-scores}
\end{equation}
 where $h_i'$ are context-dependent segment vectors computed using bi-directional GRUs (Bi-GRUs), $W_a \in \mathbb{R}^{m \times n}$ and $b_a \in \mathbb{R}^{n}$ are the attention model's weight and bias parameter, respectively, and $u_a \in \mathbb{R}^{m}$ is the ``attention query" vector parameter.
The probabilistic interpretation of Equation~\ref{eq:softmax-attention-weights} suggests that, when using the softmax attention, exactly one segment should be considered important under the constraint that the weights of all segments sum to one.
This property of the softmax attention to prioritize one instance explains the successful application of the mechanism for problems such as machine translation~\cite{bahdanau2014neural}, where the role of attention is to align each target word to (usually) one of the $M$ words from the source language.
However, softmax attention is not well suited for estimating the aggregation function weights for our problem, where multiple segments usually affect the review-level prediction.

\begin{figure}[t]
\includegraphics[height = 7.0cm, width = 8.0cm]{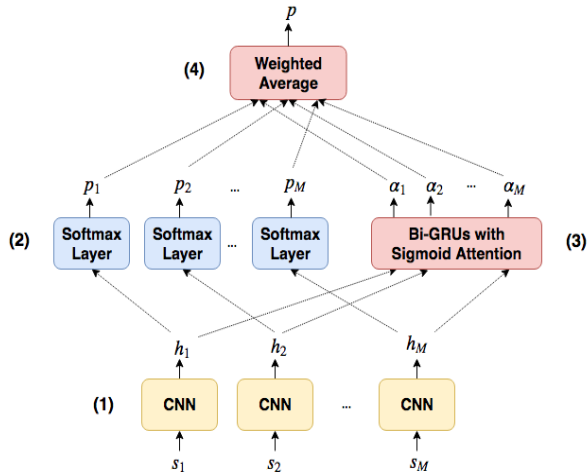}

	\caption{Our Hierarchical Sigmoid Attention Network.}
\label{fig:model_architecture}
\end{figure}

We hence propose using the sigmoid attention mechanism to compute the weights $\alpha_1,\dotsc,\alpha_M$.
In particular, we replace softmax in Equation~\eqref{eq:softmax-attention-weights} with the sigmoid (logistic) function:
\begin{equation}
     \alpha_i = \sigma(e_i) = \frac{1}{1+\exp(-e_i)}.
     \label{eq:sigmoid-attention-weights}
\end{equation}
With sigmoid attention, the computation of the attention weight $\alpha_i$ does not depend on scores $e_j$ for $j \neq i$.
Indeed, the probabilistic interpretation of sigmoid attention 
is a vector $z$ of discrete latent variables $z = [z_1, \dotsc , z_M]$, where  $z_i \in \{0, 1\}$~\cite{kim2017structured}.
In other words, the relative importance of each segment is modeled as a Bernoulli distribution.  
The sigmoid attention probability distribution is:
\begin{equation}
p(z_i = 1 \mid e_1,\dotsc,e_M) = \sigma(e_i).
\end{equation}
This probabilistic model indicates that  $z_1, \dotsc, z_M$ are conditionally independent given $e_1,\dotsc,e_M$.
Therefore, sigmoid attention allows multiple segments, or even no segments, to be selected. %
This property of sigmoid attention explains why it is more appropriate for our problem.
Also, as we will see in the next sections, using the sigmoid attention is the key modeling change needed in MIL-based hierarchical networks to outperform non-hierarchical baselines for segment-level classification. 
Attention mechanisms using sigmoid activation have also been recently applied for tasks different than segment-level classification of reviews~\cite{shen2016neural,kim2017structured,rei2018zero}.
Our work differs from these approaches in that we use the sigmoid attention mechanism for the MIL aggregation function of Equation~\ref{eq:weighted-avg-aggregation}, i.e., we aggregate segment labels $p_i$ (instead of segment vectors $h_i$) into a single review label $p$ (instead of review vectors $h$).

We summarize our HSAN architecture in Figure~\ref{fig:model_architecture}.
HSAN follows the MIL framework and thus it does not require segment labels for training. 
Instead, we only use ground-truth review labels and jointly learn the model parameters by minimizing the negative log-likelihood of the model parameters. 
Even though a single label is available for each review, our model allows different segments of the review to receive different labels. 
Thus, we can appropriately handle reviews such as that in Figure~\ref{fig:food_poisoning_example} and assign a mix of positive and negative segment labels, even when the review as a whole has a negative (2-star) rating.

\section{Experiments}
\label{s:application-segment-level-sentiment-classification}
We now turn to another key contribution of our paper, namely, the evaluation of critical aspects of hierarchical approaches and also our HSAN approach. 
For this, we focus on two important and fundamentally different, real-world applications:
segment-level sentiment classification and the discovery of foodborne illness in restaurant reviews.

\subsection{Experimental Settings}
\label{s:sentiment-classification-experimental-setting}
For segment-level sentiment classification, we use the Yelp'13 corpus with 5-star ratings~\cite{tang2015document} and the IMDB corpora with 10-star ratings~\cite{diao2014jointly}.
We do not use segment labels for training any models except the fully supervised Seg-* baselines (see below). 
For evaluating the segment-level classification performance on Yelp'13 and IMDB, we use %
the SPOT-Yelp and SPOT-IMDB datasets, respectively~\cite{angelidis2018multiple}, annotated at two levels of granularity, namely, sentences (SENT) and Elementary Discourse Units (EDUs)\footnote{The use of EDUs for sentiment classification is motivated in~\cite{angelidis2018multiple}.} (see Table~\ref{tab:spot-statistics}).
For dataset statistics and implementation details, see the supplementary material.

\begin{table}[t]
\centering
\resizebox{\columnwidth}{!}{
\begin{tabular}{|l| c c |c c|c c}
\hline
               &      \multicolumn{2}{c|}{\textbf{SPOT-Yelp}}  &  \multicolumn{2}{c|}{\textbf{SPOT-IMDB}}      \\ 
    \textbf{Statistic}           & \textbf{SENT}  & \textbf{EDU}   & \textbf{SENT}  & \textbf{EDU}   \\ \hline
\# Segments& 1,065 & 2,110 & 1,029 & 2,398 \\\hline
Positive segments (\%) & 39.9 &32.9 & 37.9 & 25.6\\ 
Neutral segments (\%)& 21.7 &34.3 & 29.2& 47.7\\
Negative segments (\%)& 38.4 & 32.8 & 32.9 &26.7 \\\hline
Witness positive (\# segs) &  7.9 & 12.1 & 6.0 & 8.5\\
Witness negative (\# segs)  & 7.3 & 11.6 & 6.6 & 11.2\\
Witness salient (\# segs)  & 8.5 & 14.0 & 7.6 & 12.6\\ \hline
\hline
WR positive  & 0.74  & 0.58 & 0.55 & 0.36\\
WR negative  & 0.68 & 0.53 & 0.63 & 0.43\\
WR salient  & 0.80 & 0.65 & 0.76 & 0.55\\
\hline
\end{tabular}}
\caption{Label statistics for the SPOT datasets. ``WR ($x$)'' is the witness rate, meaning the proportion of segments with label $x$ in a review with label $x$.
``Witness ($x$)'' is the average number of segments with label $x$ in a review with label $x$. ``Salient'' is the union of the ``positive'' and ``negative'' classes.}
\label{tab:spot-statistics}
\end{table}

For the discovery of foodborne illness, we use a dataset of Yelp restaurant reviews, manually labeled by epidemiologists in the New York City Department of Health and Mental Hygiene. 
Each review is assigned a binary label (``Sick'' vs. ``Not Sick'').
To test the models at the sentence level,
epidemiologists have manually annotated each sentence for a subset of the test reviews (see the supplementary material). In this sentence-level dataset, the WR of the ``Sick'' class is 0.25, which is significantly lower than the WR on sentiment classification datasets (Table~\ref{tab:spot-statistics}).
In other words, the proportion of ``Sick'' segments in ``Sick'' reviews is relatively low; in contrast, in sentiment classification the proportion of positive (or negative) segments is relatively high in positive (or negative) reviews.

For a robust evaluation of our approach (HSAN), we compare against state-of-the-art models and baselines: 

\begin{itemize}
    \item \textbf{Rev-*:} non-hierarchical models, trained at the review level and applied at the segment level (see Section~\ref{s:non-hierarchical-baselines}); this family includes a logistic regression classifier trained on review embeddings, computed as the element-wise average of word embeddings (``Rev-LR-EMB''), a CNN (``Rev-CNN'')~\cite{kim2014convolutional}, and a Bi-GRU with attention (``Rev-RNN'')~\cite{bahdanau2014neural}.
    For foodborne classification we also report a logistic regression classifier trained on bag-of-words review vectors (``Rev-LR-BoW''), because it is the best performing model in previous work~\cite{effland2018discovering}.
    \item \textbf{MIL-*:} MIL-based hierarchical deep learning models with different aggregation functions. 
    ``MIL-avg'' computes the review label as the average of the segment-level predictions~\cite{kotzias2015group}. ``MIL-softmax'' uses the softmax attention mechanism --this is the best performing MILNET model reported in~\cite{angelidis2018multiple} (``MILNETgt''). 
    ``MIL-sigmoid'' uses the sigmoid attention mechanism as we propose in Section~\ref{s:model-description} (HSAN model). All MIL-* models have the hierarchical structure of Figure~\ref{fig:weakly_supervised_segment_level_classification} and for comparison reasons we use the same functions for segment encoding (ENC) and segment classification (CLF), namely, a CNN and a softmax classifier, respectively. %
\end{itemize}
For the evaluation of hierarchical non-MIL networks such as the hierarchical classifier of~\citet{yang2016hierarchical}, see~\citet{angelidis2018multiple}. 
Here, we ignore this class of models as they have been outperformed by MILNET.

The above models require only review-level labels for training, which is the scenario of focus of this paper.
For comparison purposes, 
we also evaluate a family of fully supervised baselines trained at the {\em segment} level:
\begin{itemize}
    \item \textbf{Seg-*:} fully supervised baselines using SPOT segment labels for training. ``Seg-LR'' is a logistic regression classifier trained on segment embeddings, which are computed as the element-wise average of the corresponding word embeddings. We also report the CNN baseline (``Seg-CNN''), which was evaluated in~\citet{angelidis2018multiple}. Seg-* baselines are evaluated using 10-fold cross-validation on the SPOT dataset. 
\end{itemize}
For sentiment classification, we evaluate the models using the macro-averaged F1 score. 
For foodborne classification, we report both macro-averaged F1 and recall scores (for more metrics, see the supplementary material). 

\subsection{Experimental Results}
\label{s:sentiment-classification-evaluation-results}

\paragraph{Sentiment Classification:} Table~\ref{tab:spot-results} reports the evaluation results on SPOT datasets for both sentence- and EDU-level classification. 

The Seg-* baselines are not directly comparable with other models, as they are trained at the segment level on the (relatively small) SPOT datasets with segment labels.
The more complex Seg-CNN model does not significantly improve over the simpler Seg-LR, perhaps due to the small training set available at the segment level.

Rev-CNN outperforms Seg-CNN in three out of the four datasets.
Although Rev-CNN is trained at the review level (but is applied at the segment level), it is trained with 10 times as many examples as Seg-CNN.
This suggests that, for the non-hierarchical CNN models, review-level training may be advantageous with more training examples.
In addition, Rev-CNN outperforms Rev-LR-EMB, indicating that the fine-tuned features extracted by the CNN are an improvement over the pre-trained embeddings used by Rev-LR-EMB.

Rev-CNN outperforms MIL-avg and has comparable performance to MILNET: non-hierarchical deep learning models trained at the review level and applied at the segment level are strong baselines, because of their representational power. 
Thus, the Rev-* model class should be evaluated and compared with MIL-based hierarchical models for applications where segment labels are not available.

\begin{table}[t]
\resizebox{\columnwidth}{!}{
\begin{tabular}{|l|cc|cc|}
\hline
           &      \multicolumn{2}{c|}{\textbf{SPOT-Yelp}}  &  \multicolumn{2}{c|}{\textbf{SPOT-IMDB}}      \\ 
\textbf{Method}             & \textbf{SENT}  & \textbf{EDU}   & \textbf{SENT}  & \textbf{EDU}   \\ \hline
Seg-LR & 55.6& 59.2  & 60.5 & 62.8 \\ 
Seg-CNN            & 56.2 & 60.0 & 58.3 & 63.0 \\ \hline
Rev-LR-EMB         & 51.2    & 49.3 & 52.7& 48.6 \\
Rev-CNN            & 60.6 & 61.5 & 60.8 & 60.1 \\
Rev-RNN            & 58.5 & 53.9 & 55.3 & 50.8 \\\hline 
MIL-avg           & 51.8 & 46.8 & 45.7 & 38.4 \\
MIL-softmax & 63.4 & 59.9 & 64.0 & 59.9 \\
MIL-sigmoid  & \textbf{64.6} & \textbf{63.3} & \textbf{66.2} & \textbf{65.7}  \\\hline

\hline 
\end{tabular}}
\caption{F1 score for segment-level sentiment classification.}
\label{tab:spot-results}
\end{table}

Interestingly, MIL-sigmoid (HSAN) consistently outperforms all models, including MIL-avg, MIL-softmax (MILNET), and the Rev-* baselines. 
This shows that:
\begin{enumerate}
\item the choice of aggregation function of MIL-based classifiers heavily impacts classification performance; and 
\item MIL-based hierarchical networks can indeed outperform non-hierarchical networks when the appropriate aggregation function is used.
\end{enumerate}
We emphasize that we use the same ENC and CLF functions across all MIL-based models to show that performance gains stem solely from the choice of aggregation function.
Given that HSAN consistently outperforms MILNET in all datasets for segment-level sentiment classification, we conclude that the choice of sigmoid attention for aggregation is a better fit than softmax for this task. 

The difference in performance between HSAN and MILNET is especially pronounced on the *-EDU datasets.
We explain this behavior with the statistics of Table~\ref{tab:spot-statistics}: ``Witness (Salient)'' is higher in *-EDU datasets compared to *-SENT datasets. 
In other words, *-EDU datasets contain more segments that should be considered important than *-SENT datasets.
This implies that the attention model needs to ``attend'' to more segments in the case of *-EDU datasets: as we argued in Section~\ref{s:model-description}, this is best modeled by sigmoid attention.

\paragraph{Foodborne Illness Discovery:} Table~\ref{tab:foodborne-eval-results} reports the evaluation results for both review- and sentence-level foodborne classification.\footnote{We report review-level classification results because epidemiologists rely on the {\em review-level} predictions to decide whether to investigate restaurants; in turn, {\em segment-level} predictions help epidemiologists focus on the relevant portions of positively labeled reviews.}
For more detailed results, see the supplementary material.
Rev-LR-EMB has significantly lower F1 score than Rev-CNN and Rev-RNN: representing a review as the uniform average of the word embeddings is not an appropriate modeling choice for this task, where only a few segments in each review are relevant to the positive class. 

MIL-sigmoid (HSAN) achieves the highest F1 score among all models for review-level classification. 
MIL-avg has lower F1 score compared to other models: as discussed in Section~\ref{s:multiple-instance-learning}, in applications where the value of WR is very low (here WR=0.25), the uniform average is not an appropriate aggregation function for MIL.

Applying the best classifier reported in~\citet{effland2018discovering} (Rev-LR-BoW) for sentence-level classification leads to high precision but very low recall. 
On the other hand, the MIL-* models outperform the Rev-* models in F1 score (with the exception of MIL-avg, which has lower F1 score than Rev-RNN): the MIL framework is appropriate for this task, especially when the weighted average is used for the aggregation function. 
The significant difference in recall and F1 score between different MIL-based models highlights once again the importance of choosing the appropriate aggregation function. 
MIL-sigmoid consistently outperforms MIL-softmax in all metrics, showing that the sigmoid attention properly encodes the hierarchical structure of reviews.
MIL-sigmoid also outperforms all other models in all metrics. 
Also, MIL-sigmoid's recall is 48.6\% higher than that of Rev-LR-BoW. 
In other words, MIL-sigmoid detects more sentences relevant to foodborne illness than Rev-LR-BoW, which is especially desirable for this application, as discussed next. 

\begin{table}[t]
\resizebox{\columnwidth}{!}{
\begin{tabular}{|l|c|cccc|}
\hline
           &      \multicolumn{1}{c|}{\textbf{REV}}  &  \multicolumn{4}{c|}{\textbf{SENT}}      \\ 
\textbf{Method}  &  \textbf{F1} & \textbf{Prec}   & \textbf{Rec}  & \textbf{F1} & \textbf{AUPR}   \\ \hline
Rev-LR-BoW & 86.7 & \textbf{82.1} & 58.8 & 68.6 & 80.9  \\ 
Rev-LR-EMB    & 63.3     & 50.0 & 84.3  & 62.8    & 48.9 \\
Rev-CNN   & 84.8          & 79.3 & 59.4 & 67.9 & 24.7 \\
Rev-RNN      & 86.7       &81.0 & 74.5 & 77.6 & 11.3  \\\hline 
MIL-avg     & 59.8      & 75.0 & 78.0  & 76.5 & 73.6 \\
MIL-softmax & 87.6 & 75.5 & 83.3 & 79.2 & 81.6 \\
MIL-sigmoid  & \textbf{89.6} & 76.4 & \textbf{87.4}  & \textbf{81.5} & \textbf{84.0} \\\hline
\end{tabular}}
\caption{Review-level (left) and sentence-level (right) evaluation results for discovering foodborne illness.}
\label{tab:foodborne-eval-results}
\end{table}

\begin{figure}[t]
\includegraphics[height = 6.2cm, width = 8.0cm]{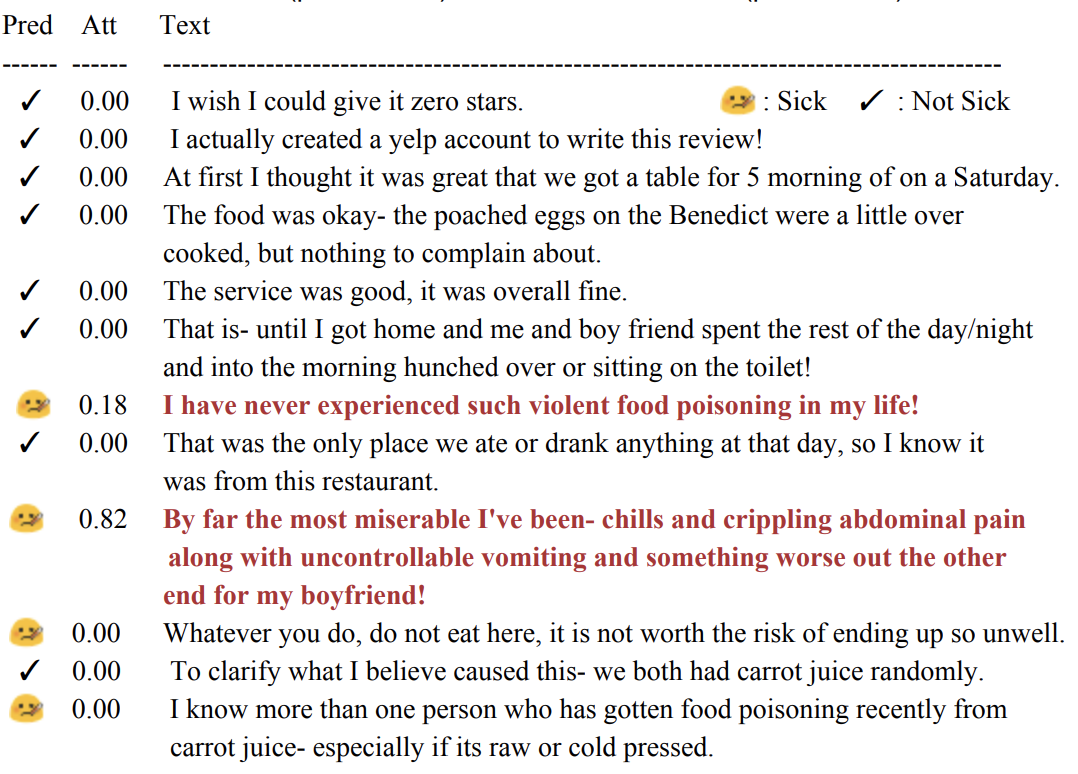}
	\caption{HSAN's fine-grained predictions for a Yelp review: for each sentence, HSAN provides one binary label (Pred) and one attention score (Att). A sentence is highlighted if its attention score is greater than $0.1$.}
\label{fig:wshan_demo}
\end{figure}

\paragraph{Important Segment Highlighting}
\label{s: importante-segment-highlighting}
Fine-grained predictions could potentially help epidemiologists to quickly focus on the relevant portions of the reviews and safely ignore the rest. Figure~\ref{fig:wshan_demo} shows how the segment predictions and attention scores predicted by HSAN ---with the highest recall and F1 score among all models that we evaluated--- could be used to highlight important sentences of a review.
We highlight sentences in red if the corresponding attention scores exceed a pre-defined threshold. 
In this example, high attention scores are assigned by HSAN to sentences that mention food poisoning or symptoms related to food poisoning. 
(For more examples, see the supplementary material.)
This is particularly important because reviews on Yelp and other platforms can be long, with many irrelevant sentences surrounding the truly important ones for the task at hand.
The fine-grained predictions produced by our model could inform a graphical user interface in health departments for the inspection of candidate reviews.
Such an interface would allow epidemiologists to examine reviews more efficiently and, ultimately, more effectively.

\section{Conclusions and Future Work}
\label{s:conclusion}
We presented a Multiple Instance Learning-based model for fine-grained text classification that requires only review-level labels for training but produces both review- and segment-level labels. 
Our first contribution is the observation that non-hierarchical deep networks trained at the review level and applied at the segment level (by treating each test segment as if it were a short ``review'') are surprisingly strong and perform comparably or better than MIL-based hierarchical networks with a variety of aggregation functions. 
Our second contribution is a new MIL aggregation function based on the sigmoid attention mechanism, which explicitly allows multiple segments to contribute to the review-level classification decision with different weights.
We experimentally showed that the sigmoid attention is the key modeling change needed for MIL-based hierarchical networks to outperform the non-hierarchical baselines for segment-level sentiment classification. 
Our third contribution is the application of our weakly supervised approach to the important public health application of foodborne illness discovery in online restaurant reviews. 
We showed that our MIL-based approach has a higher chance than all previous models to identify unknown foodborne outbreaks, and demonstrated how its fine-grained segment annotations can be used to highlight the segments that were considered important for the computation of the review-level label. 

In future work, we plan to consider alternative techniques for segment encoding (ENC), such as pre-trained transformer-based language models~\cite{devlin2019bert,radford2018improving}, which we expect to further boost our method's performance. 
We also plan to quantitatively evaluate the extent to which the fine-grained predictions of our model help epidemiologists to efficiently examine candidate reviews and to interpret classification decisions.  
Indeed, choosing segments of the review text that explain the review-level decisions can help interpretability~\cite{lei2016rationalizing,yessenalina2010multi,biran2017explanation}. 
Another important direction for future work is to study if minimal supervision at the fine-grain level, either in the form of expert labels or rationales~\cite{bao2018deriving}, could effectively guide the weakly supervised models.
This kind of supervision is especially desirable to satisfy prior beliefs about the intended role of fine-grained predictions in downstream applications.
We believe that building this kind of fine-grained models is particularly desirable when model predictions are used by humans to take concrete actions in the real world. 

\subsubsection*{Acknowledgments}
We thank the anonymous reviewers for their constructive feedback. This material is based upon work supported by the National Science Foundation under Grant No. IIS-15-63785.

\bibliography{myreferences}
\bibliographystyle{acl_natbib}

\newpage

\appendix
\section{Appendix}
\label{s:appendix}

\subsection{Sentiment Classification}
In this section, for reproducibility, we discuss all details of the datasets (Section~\ref{appendix:sentiment-classification-dataset}) as well as the configuration of the techniques and the evaluation methodology (Section~\ref{appendix:sentiment-classification-implementation-details}) for the sentiment classification experiments.

\subsubsection{Datasets}
\label{appendix:sentiment-classification-dataset}
The Yelp'13 corpus~\cite{tang2015document} contains 335,018 user reviews of local businesses. 
Each review includes a 5-star rating ranging from 1 (negative) to 5 stars (positive). 
The IMDB corpus~\cite{diao2014jointly} contains 348,415 movie reviews with ratings ranging from 1 (negative) to 10 stars (positive). 
For both corpora, training (80\%), validation (10\%), and test (10\%) sets are provided. 

For evaluation, we use the SPOT-Yelp and SPOT-IMDB datasets.
These datasets contain 100 Yelp reviews and 97 IMDB reviews from the Yelp'13 and IMDB test sets, respectively. 
Each dataset has been segmented both at sentences (SPOT-*-SENT) and EDUs (SPOT-*-EDU).
The test sets have 3 labels (Table~1): ``negative,'' ``neutral,'' and ``positive.''
For more statistics, see Tables 1 and 2 in reference~\cite{angelidis2018multiple}, as well as Table~1 in this paper.

\subsubsection{Implementation Details}
\label{appendix:sentiment-classification-implementation-details}

\paragraph{Model Parameters}
\label{appendix:model-parameters}
For a fair comparison, all the MIL-* models have the same parameter configuration as MILNET (Section 5.3 in ~\citet{angelidis2018multiple}). 
For all models using word embeddings (i.e., Seg-*, Rev-*, MIL-*), we initialize the word embeddings using 300-dimensional ($k=300$) pre-trained word2vec embeddings~\cite{mikolov2013distributed}.
For the CNNs we use kernels of size 3, 4, and 5 words, 100 feature maps per kernel, stride of size 1, and max-over-time pooling to get fixed-size segment encodings (resulting in $\ell=300$).
For the forward and backward GRUs we use hidden vectors with 50 dimensions ($n=2 \cdot 50 = 100$), while for the attention mechanism we use vectors of 100 dimensions ($m=100$).
We use dropout (with rate 0.5) on the word embeddings and the internal GRU states. 
We use L2 regularization for the softmax classifier.%

\paragraph{Training and Validation Procedure}
We segment the training and validation reviews into sentences\footnote{We do not segment the reviews into EDUs, because this procedure requires the use of a Rhetorical Structure Theory parser, which does not exist for every language. 
Instead, we opt for a language independent model. 
At test time, the same model is applied on both sentences and EDUs.} and use the available review labels for training our model, over 5 classes for Yelp'13 and 10 classes for IMDB.
We group the training reviews in mini-batches of 200 reviews so that reviews under the same mini-batch have a similar number of segments $M$.
Thus, we allow for training the models using different values of $M$ per batch while at the same time we minimize the amount of zero-padding, leading to more efficient training. 
As an objective function, we use the negative log-likelihood of the model parameters.
We train our models using the Adadelta optimizer~\cite{zeiler2012adadelta} (with learning rate 0.005) for up to 50 epochs and we stop the training process if the validation loss does not decrease for more than 10 epochs. 
We fine-tune the model parameters on the validation set. 

\begin{table*}[!htb]
\renewcommand\thetable{4}
\centering
\resizebox{2\columnwidth}{!}{
\begin{tabular}{|l|c c c c| c c c c c|}
\hline
               &      \multicolumn{4}{c|}{\textbf{Review-Level Evaluation}}  &  \multicolumn{5}{c|}{\textbf{Sentence-Level Evaluation}}      \\ 

\textbf{Model} & \textbf{Prec} & \textbf{Rec} & \textbf{F1 (95\% CI)} &  \textbf{AUPR (95\% CI)} &  \textbf{Acc} &\textbf{Prec} & \textbf{Rec} & \textbf{F1}  & \textbf{AUPR} \\ 
\hline
KWRD1 & 0.801 & 0.581 & 0.673  (0.646, 0.699) & 0.194 (0.179, 0.208) & 0.850 & 0.806 & 0.342 & 0.481 & 0.408 \\
KWRD2 & 0.532 & 0.898 & 0.668  (0.647, 0.689) & 0.033  (0.027, 0.040) & 0.890 & 0.778 & 0.640 & 0.703 & 0.572\\ \hline
Rev-LR-BoW& 0.853 & 0.882 & 0.867  (0.852, 0.882) & 0.914  (0.900, 0.929)  & 0.891 & \textbf{0.821} & 0.588 & 0.685 & 0.809\\ 
Rev-LR-EMB &  0.704 & 0.574& 0.633 (0.513, 0.714) &	0.696  (0.649, 0.755)  & 0.797 & 0.500 & 0.843 & 0.628 & 0.489\\ 
Rev-CNN &  0.803 & 0.898 & 0.848  (0.832, 0.866)  & 0.935   (0.923, 0.946)  & 0.887 & 0.793 & 0.594 & 0.679 & 0.247\\ 
Rev-RNN &  0.856 & 0.878& 0.867  (0.849, 0.884) & 0.929  (0.915, 0.942)  & 0.913 & 0.810 & 0.745 & 0.776 & 0.113\\ \hline
MIL-avg &  0.674  & 0.537 & 0.598 (0.485, 0.682) & 0.643 (0.596, 0.708)  & 0.903 & 0.750 & 0.780 & 0.765 & 0.736\\
MIL-softmax&   0.829 & 0.928 & 0.876   (0.859, 0.890)&  \textbf{0.941}  (0.926, 0.994)  & 0.912 & 0.755 & 0.833 & 0.792 & 0.816\\
MIL-sigmoid & \textbf{0.865} & \textbf{0.929} & \textbf{0.896} (0.882, 0.910) & 0.913  (0.887, 0.926) & \textbf{0.920} & 0.764 & \textbf{0.874} & \textbf{0.815} & \textbf{0.840}\\ \hline
\end{tabular}}
\caption{Review-level (left) and sentence-level (right) evaluation results for discovering foodborne illness in Yelp reviews.}
\label{fig:foodborne-eval-results-detailed}
\end{table*}

\paragraph{Evaluation Procedure}
While the training and validation sets have 5 labels, for Yelp'13, and 10 labels, for IMDB, the test sets have 3 labels. 
During evaluation, we address this discrepancy by following the same procedure as in~\citet{angelidis2018multiple} to map the segment probability distributions from 5 classes--for Yelp'13--and 10 classes---for IMDB--to 3 classes, namely, ``negative,'' ``neutral,'' and ``positive'':
\begin{enumerate}
    \item We map the predicted probability distribution $p_i$ for each segment $s_i$ into a polarity score %
$g_{s_i} = \sum_c p_i^{c} \cdot w^{c} \in [-1, 1]$, where $w=\langle w^1,\dotsc,w^C | w^c \in [-1,1] \rangle$. The weights $w_c$ are spaced uniformly such that $w^{c+1}-w^c = \frac{2}{C-1}$. In particular, for the 5-class setting (Yelp) we get: $w = \langle -1, -0.5, 0, 0.5, 1 \rangle$, while for the 10-class setting (IMDB) we get:  $w = \langle -1, -0.778, -0.556, -0.333,$ $-0.111, 0.111, 0.333, 0.556, 0.778, 1 \rangle$.
\item We compute a gated polarity score $g'_{s_i} = \alpha_i \cdot g_{s_i}$, where $\alpha_i$ is the attention weight assigned to $s_i$ by the model.
\item We map each score $g'_{s_i}$ to one of the 3 discrete labels using two thresholds $t_1$, $t_2 \in [-1, 1]$: segment $s_i$ is classified as ``negative'' if $g'_{s_i}<t_1$, ``positive'' if $g'_{s_i}>t_2$, and ``neutral'' otherwise. 
\end{enumerate}
We evaluate the models using the macro-averaged F1 score. 
We determine the value of the $t_1$ and $t_2$ thresholds using 10-fold cross-validation and report the mean scores across the 10 folds. 

\subsection{Discovering Foodborne Illness}
In this section, for reproducibility, we discuss all details of the datasets (Section~\ref{appendix-foodborne-datasets}) as well as the configuration of the techniques and the evaluation methodology (Section~\ref{appendix:foodborne-implementation-details}) for the experiments regarding the foodborne application.

\subsubsection{Datasets}
\label{appendix-foodborne-datasets}
We use the same training and test sets as in~\cite{effland2018discovering}. 
The review-level training set (``Silver'' set in~\cite{effland2018discovering}) contains 21,551 (5,895 ``Sick,'' 15,656 ``Not Sick'') reviews posted before January 1, 2017. 
The review-level test set contains 2,975 (949 ``Sick,'' 2,026 ``Not Sick'') reviews posted after January 1, 2017. 
Sample weights are also calculated to account for the selection bias in this dataset~\cite{effland2018discovering}.

To test the ability of the models to detect sentences of the ``Sick'' reviews discussing food poisoning, epidemiologists annotated each sentence for 437 out of the 949 ``Sick" test reviews.
Given a review for labeling, epidemiologists read the whole review text and decided on the label for each sentence. 
This led to 3,114 labeled sentences (630 ``Sick,'' 2,484 ``Not Sick'').
For this application, EDU-level labels were not available, so we consider only sentences as review segments.

\subsubsection{Implementation Details}
\label{appendix:foodborne-implementation-details}
\paragraph{Model Parameters}
For the *-BoW classifiers, the review text is encoded as a bag-of-words vector including n-grams (for n=1, 2, and 3) and each term is weighted using the Term Frequency-Inverse Document Frequency (TF-IDF) statistic~\cite{leskovec2014mining}.
For the Rev-* and MIL-* classifiers, we use the same model parameter configuration as in Section~\ref{appendix:model-parameters}.
We initialize the word embeddings using 300-dimensional pre-trained word2vec embeddings.

\paragraph{Training and Validation Procedure}
We split the review-level training set into training (90\%) and validation (10\%) sets, randomly stratified by label and sample weight. 
We do not use any sentence-level labels for training. 
We group the training reviews in mini-batches of 200 reviews so that reviews under the same mini-batch have a similar number of segments. 
We train our models using the Adadelta optimizer for up to 50 epochs and we stop the training process if the validation loss does not decrease for more than 10 epochs.
We fine-tune the model parameters on the validation set with respect to the F1 score.

\paragraph{Evaluation Procedure}
Given a test review, we predict a label for each sentence and aggregate the sentence predictions to get a single review prediction. 
For review-level classification, we use the review prediction, while for sentence-level evaluation we use the individual sentence predictions. 
The segment-level confidence scores are computed by multiplying the segment probability for the ``Sick'' class with its attention weight. 
To account for the selection bias in the review-level test set, we compute precision and recall using sample weights~\cite{effland2018discovering}.
Because of the class imbalance at both the review and sentence levels, we report precision, recall, F1 score, and area under the precision-recall curve (AUPR).
Also, we follow~\citet{effland2018discovering} and estimate 95\% confidence intervals (95\% CI) for the F1 and AUPR metrics using the percentile bootstrap method~\cite{efron1994introduction} with sampled test sets of 1,000 reviews. 
For sentence-level classification, we also report the accuracy score.

\subsubsection{More Results and Examples}
\label{appendix:foodborne-implementation-details}

\paragraph{Detailed Evaluation Results}
Table~\ref{fig:foodborne-eval-results-detailed} includes the evaluation results, which were reported in Table~3, as well as more baselines and evaluation metrics. 
For completeness, we also evaluate the ``KWRD*'' class of keyword search classifiers: ``KWRD1'' predicts the ``Sick'' class if the ``food poisoning'' phrase is included in the (lemmatized and lower cased) review text. ``KWRD2'' predicts the ``Sick'' class if at least one of the following terms are included in the review text: ``food poisoning,'' ``sick,'' ``vomit,'' ``diarrhea.'' 

\end{document}